\documentclass[letterpaper]{article} 
\usepackage{aaai23}  
\usepackage{times}  
\usepackage{helvet}  
\usepackage{courier}  
\usepackage[hyphens]{url}  
\usepackage{graphicx} 
\usepackage{natbib}  
\usepackage{caption} 
\usepackage[utf8]{inputenc}
\usepackage[T1]{fontenc}

\usepackage{newfloat}
\usepackage{listings}
\usepackage{hyperref}       
\usepackage{url}            
\usepackage{booktabs}       
\usepackage{amsfonts}       
\usepackage{nicefrac}       
\usepackage{microtype}      
\usepackage{xcolor}         
\usepackage{multicol}
\usepackage{multirow}
\usepackage{amsmath}
\usepackage{amsfonts}
\usepackage{amssymb}
\usepackage{amsthm}
\usepackage{breqn}
\usepackage{stmaryrd}
\usepackage{algorithm}
\usepackage{algpseudocode}
\usepackage{placeins}
\usepackage{dsfont}
\newcommand{\sem}[1]{\llbracket #1 \rrbracket}
\newcommand{\semink}[2]{\sem{#1}_{\text{int-}#2}}
\newcommand{\semfloat}[1]{\sem{#1}_{\text{float32}}}

\newtheorem{theorem}{Theorem}
\newtheorem{corollary}{Corollary}

\newcommand{\Ind}{\mathds{1}}

\DeclareCaptionStyle{ruled}{labelfont=normalfont,labelsep=colon,strut=off} 
\floatstyle{ruled}
\newfloat{listing}{tb}{lst}{}
\floatname{listing}{Listing}
\title{Quantization-aware Interval Bound Propagation for Training Certifiably Robust Quantized Neural Networks}
\author{
Mathias Lechner\textsuperscript{\rm 1}, \DJ{}or\dj{}e \v{Z}ikeli\'c\textsuperscript{\rm 2}, Krishnendu Chatterjee\textsuperscript{\rm 2}, Thomas A. Henzinger\textsuperscript{\rm 2}, Daniela Rus\textsuperscript{\rm 1}
}
\affiliations{
    \textsuperscript{\rm 1}Massachusetts Institute of Technology (MIT)\\
    \textsuperscript{\rm 2}Institute of Science and Technology Austria (ISTA)\\

    Correspondence to mlechner@mit.edu
}

\usepackage{bibentry}

\begin{document}

\maketitle

\begin{abstract}
We study the problem of training and certifying adversarially robust quantized neural networks (QNNs). Quantization is a technique for making neural networks more efficient by running them using low-bit integer arithmetic and is therefore commonly adopted in industry. Recent work has shown that floating-point neural networks that have been verified to be robust can become vulnerable to adversarial attacks after quantization, and certification of the quantized representation is necessary to guarantee robustness.
In this work, we present quantization-aware interval bound propagation (QA-IBP), a novel method for training robust QNNs.
Inspired by advances in robust learning of non-quantized networks, our training algorithm computes the gradient of an abstract representation of the actual network. Unlike existing approaches, our method can handle the discrete semantics of QNNs. 
Based on QA-IBP, we also develop a complete verification procedure for verifying the adversarial robustness of QNNs, which is guaranteed to terminate and produce a correct answer. Compared to existing approaches, the key advantage of our verification procedure is that it runs entirely on GPU or other accelerator devices. 
We demonstrate experimentally that our approach significantly outperforms existing methods and establish the new state-of-the-art for training and certifying the robustness of QNNs.
\end{abstract}

\section{Introduction}
Quantized neural networks (QNNs) are neural networks that represent their weights and compute their activations using low-bit integer variables. QNNs significantly improve the latency and computational efficiency of inferencing the network for two reasons. 
First, the reduced size of the weights and activations allows for a much more efficient use of memory bandwidth and caches. Second, integer arithmetic requires less silicon area and less energy to execute than floating-point operations. Consequently, dedicated hardware for running QNNs can be found in GPUs, mobile phones, and autonomous driving computers. 

Adversarial attacks are a well-known vulnerability of neural networks that raise concerns about their use in safety-critical applications \cite{szegedy2013intriguing,goodfellow2014explaining}. These attacks are norm-bounded input perturbations that make the network misclassify samples, despite the original samples being classified correctly and the perturbations being barely noticeable by humans. For example, most modern image classification networks can be fooled when changing each pixel value of the input image by a few percent.
Consequently, researchers have tried to train networks that are provably robust against such attacks.
The two most common paradigms of training robust networks are adversarial training \cite{madry2017towards}, and abstract interpretation-based training \cite{mirman2018differentiable,wong2018provable}.
Adversarial training and its variations perturb the training samples with gradient descent-based adversarial attacks before feeding them into the network \cite{madry2017towards,zhang2019theoretically,wu2020adversarial,lechner2021adversarial}. While this improves the robustness of the trained network empirically, it provides no formal guarantees of the network's robustness due to the incompleteness of gradient descent-based attacking methods, i.e., gradient descent might not find all attacks.
Abstract interpretation-based methods avoid this problem by overapproximating the behavior of the network in a forward pass during training. In particular, instead of directly training the network by computing gradients with respect to concrete samples, these algorithms compute gradients of bounds obtained by propagating abstract domains. 
While the learning process of abstract interpretation-based training is much less stable than a standard training procedure, it provides formal guarantees about the network's robustness. The interval bound propagation (IBP) method \cite{gowal2019scalable} effectively showed that the learning process with abstract interpretation can be stabilized when gradually increasing the size of the abstract domains throughout the training process.

Previous work has considered adversarial training and IBP for floating-point arithmetic neural networks, however robustness of QNNs has received comparatively much less attention. Since it was demonstrated by~\cite{giacobbe2020many} that neural networks may become vulnerable to adversarial attacks after quantization even if they have been verified to be robust prior to quantization, one must develop specialized training and verification procedures in order to guarantee robustness of QNNs. Previous works have proposed several robustness {\em verification} procedures for QNNs~\cite{giacobbe2020many,baranowski2020smt,henzinger2021scalable}, but none of them consider algorithms for {\em learning} certifiably robust QNNs. Furthermore, the existing verification procedures are based on constraint solving and cannot be run on GPU or other accelerating devices, making it much more challenging to use them for verifying large QNNs.

In this work, we present the first abstract interpretation training method for the discrete semantics of QNNs. We achieve this by first defining abstract interval arithmetic semantics that soundly over-approximate the discrete QNN semantics, giving rise to an end-to-end differentiable representation of a QNN abstraction. We then instantiate quantization-aware training techniques within these abstract interval arithmetic semantics in order to obtain a procedure for training certifiably robust QNNs.

Next, we develop a robustness verification procedure which allows us to formally verify QNNs learned via our IBP-based training procedure. We prove that our verification procedure is {\em complete}, meaning that for any input QNN it will either prove its robustness or find a counterexample. This contrasts the case of abstract interpretation verification procedures for neural networks operating over real arithmetic which are known to be incomplete~\cite{mirman2021fundamental}. The key advantage of our training and verification procedures for QNNs is that it can make use of GPUs or other accelerator devices. In contrast, the existing verification methods for QNNs are based on constraint solving so cannot be run on GPU.

Finally, we perform an experimental evaluation showing that our method outperforms existing state-of-the-art certified $L_\infty$-robust QNNs.
We also elaborate on the limitations of our training method by highlighting how the low precision of QNNs makes IBP-based training difficult.


We summarize our contribution in three points:
\begin{itemize}
    \item We introduce the first learning procedure for learning robust QNNs. Our learning procedure is based on quantization-aware training and abstract interpretation method.
    \item We develop the first \emph{complete} robustness verification algorithm for QNNs (i.e., one that always terminates with the correct answer) that runs entirely on GPU or other neural network accelerator devices and make it publicly available.
    \item We experimentally demonstrate that our method advances the state-of-the-art on certifying $L_\infty$-robustness of QNNs.
\end{itemize}

\section{Related Work}

\paragraph{Abstract interpretations for neural networks}
Abstract interpretation is a method for overapproximating the semantics of a computer program in order to make its formal analysis feasible~\cite{cousot1977abstract}. Abstract interpretation executes program semantics over abstract domains instead of concrete program states. The method has been adapted to the robustness certification of neural networks by computing bounds on the outputs of neural networks \cite{wong2018provable, GehrMDTCV18,tjeng2019evaluating}.
For instance, polyhedra \cite{,katz2017reluplex,ehlers2017formal,GehrMDTCV18,singh2019abstract, tjeng2019evaluating}, intervals \cite{gowal2019scalable}
, hybrid automata \cite{xiang2018output}, zonotopes \cite{singh2018fast}, convex relaxations \cite{dvijotham2018dual,ZhangCXGSLBH20,wang2021beta}, and polynomials \cite{zhang2018efficient} have been used as abstract domains in the context of neural network verification.
Abstract interpretation has been shown to be most effective for verifying neural networks when directly incorporating them into gradient descent-based training algorithms by optimizing the obtained output bounds as the loss function \cite{mirman2018differentiable,wong2018provable,gowal2019scalable,ZhangCXGSLBH20}.

Most of the abstract domains discussed above exploit the piecewise linear structure of neural networks, e.g., linear relaxations such as polytopes and zonotopes. However, linear relaxations are less suited for QNNs due to their piecewise-constant discrete semantics. 

\paragraph{Verification of quantized neural networks}
The earliest work on the verification of QNNs has focused on binarized neural networks (BNNs), i.e., 1-bit QNNs \cite{hubara2016binarized}. In particular, \cite{narodytska2018verifying} and \cite{cheng2018verification} have reduced the problem of BNN verification to boolean satisfiability (SAT) instances. Using modern SAT solvers, the authors were able to verify formal properties of BNNs.
\citep{jia2020efficient} further improve the scalability of BNNs by specifically training networks that can be handled by SAT-solvers more efficiently.
\citep{amir2021smt} developed a satisfiability modulo theories (SMT) approach for BNN verification.

Verification for many bit QNNs was first reported in \citep{giacobbe2020many} by reducing the QNN verification problem to quantifier-free bit-vector satisfiability modulo theory (QF\_BV SMT). The SMT encoding was further improved in \cite{henzinger2021scalable} by removing redundancies from the SMT formulation.
\citep{baranowski2020smt} introduced fixed-point arithmetic SMT to verify QNNs. The works of \cite{sena2021verifying,sena2022automated} have studied SMT-based verification for QNNs as well. Recently, \cite{mistry2022milp} proposed encoding of the QNN verification problem into a mixed-integer linear programming (MILP) instance.
IntRS \cite{lin2021integer} considers the problem of certifying adversarial robustness of quantized neural networks using randomized smoothing. IntRS is limited to $L_2$-norm bounded attacks and only provides statistical instead of formal guarantees compared to our approach.  

\paragraph{Decision procedures for neural network verification}
Early work on the verification of floating-point neural networks has employed off-the-shelf tools and solvers. For instance, \cite{pulina2012challenging,katz2017reluplex,ehlers2017formal} employed SMT-solvers to verify formal properties of neural networks. Similarly, \cite{tjeng2019evaluating} reduces the verification problem to mixed-integer linear programming instances.
Procedures better tailored to neural networks are based on branch and bound algorithms \cite{bunel2018unified}. In particular, these algorithms combine incomplete verification routines (bound) with divide-and-conquer (branch) methods to tackle the verification problem.
The speedup advantage of these methods comes from the fact that the bounding methods can be easily implemented on GPU and other accelerator devices \cite{gpupoly,wang2021beta}. 

\paragraph{Quantization-aware training} There are two main strategies for training QNNs: post-training quantization and quantization-aware training \cite{krishnamoorthi2018quantizing}. In post-training quantization, a standard neural network is first trained using floating-point arithmetic, which is then translated to a quantized representation by finding suitable fixed-point format that makes the quantized interpretation as close to the original network as possible. Post-training quantization usually results in a drop in the accuracy of the network with a magnitude that depends on the specific dataset and network architecture.

To avoid a significant reduction in accuracy caused by the quantization in some cases, {\em quantization-aware training (QAT)} models the imprecision of the low-bit fixed-point arithmetic already during the training process, i.e., the network can adapt to a quantized computation during training.
The rounding operations found in the semantics of QNNs are non-differentiable computations. Consequently, QNNs cannot be directly trained with stochastic gradient descent. 
Researchers have come up with several ways of circumventing the problem of non-differentiable rounding.
The most common approach is the straight-through gradient estimator (STE) \cite{bengio2013estimating,hubara2017quantized}. In the forward pass of a training step, the STE applies rounding operations to computations involved in the QNN, i.e., the weights, biases, and arithmetic operations. However, in the backward pass, the rounding operations are removed such that the error can backpropagate through the network. 
The approach of \citep{gupta2015deep} uses stochastic rounding that randomly selects one of the two nearest quantized values for a given floating-point value. 
Relaxed quantization \cite{louizos2019relaxed} generalizes stochastic rounding by replacing the probability distribution over the nearest two values with a distribution over all possible quantized values. 
DoReFA-Net \cite{zhou2016dorefa} combines the straight-through gradient estimator and stochastic rounding to train QNN with high accuracy. The authors observed that quantizing the first and last layer results in a significant drop in accuracy, and therefore abstain from quantizing these two layers.

Instead of having a fixed pre-defined quantization range, i.e., fixed-point format, more recent QAT schemes allow learning the quantization range. 
PACT \cite{choi2018pact} treats the maximum representable fixed-point value as a free variable that is learned via stochastic gradient descent using a straight-through gradient estimation.
The approach of \citep{jacob2018quantization} keeps a moving average of the values ranges during training and adapts the quantization range according to the moving average.
LQ-Nets \cite{zhang2018lq} learn an arbitrary set of quantization levels in the form of a set of coding vectors. While this approach provides a better approximation of the real-valued neural network than fixed-point-based quantization formats, it also prevents the use of efficient integer arithmetic to run the network.
MobileNet \cite{howard2019searching} is a specialized network architecture family for efficient inference on the ImageNet dataset and employs quantization as one technique to achieve this target.
HAWQ-V3 \cite{yao2021hawq} dynamically assigns the number of bits of each layer to either 4-bit, 8-bit, or 32-bit depending on how numerically sensitive the layer is.
EfficientNet-lite \cite{tan2019efficientnet} employs a neural architecture search to automatically find a network architecture that achieves high accuracy on the ImageNet dataset while being fast for inference on a CPU.

\section{Preliminaries}\label{sec:prelims}

\paragraph{Quantized neural networks (QNNs)} Feedforward neural networks are functions $f_{\theta}: \mathbb{R}^n\rightarrow\mathbb{R}^m$ that consist of several sequentially composed layers $f_{\theta} = l_1\circ\dots\circ l_s$, where layers are parametrized by the vector $\theta$ of neural network parameters. Quantization is an interpretation of a neural network $f_{\theta}$ that evaluates the network over a fixed point arithmetic and operates over a restricted set of bitvector inputs~\citep{smith1997scientist}, e.g.~$4$ or $8$ bits. Formally, given an admissible input set $\mathcal{I}\subseteq \mathcal{R}^n$, we define an interpretation map
\[ \sem{\cdot}_{\mathcal{I}}: (\mathbb{R}^n\rightarrow\mathbb{R}^m) \rightarrow (\mathcal{I}\rightarrow\mathbb{R}^m), \]
which maps a neural network to its interpretation operating over the input set $\mathcal{I}$. For instance, if $\mathcal{I}=\mathbb{R}^n$ then $\sem{f_{\theta}}_{\mathbb{R}}$ is the idealized real arithmetic interpretation of $f_{\theta}$, whereas $\semfloat{f}$ denotes its floating-point 32-bit implementation~\citep{kahan1996ieee}. Given $k\in\mathbb{N}$, the k-bit {\em quantization} is then an interpretation map $\semink{\cdot}{k}$ which uses $k$-bit fixed-point arithmetic. We say that $\semink{f_{\theta}}{k}$ is a $k$-bit {\em quantized neural network (QNN)}.

The semantics of the QNN $\semink{f_{\theta}}{k}$ are defined as follows. Let $[\mathbb{Z}]_k=\{0,1\}^k$ be the set of all bit-vectors of bit-width $k$. The QNN $\semink{f_{\theta}}{k}$ then also consists of sequentially composed layers $\semink{f_{\theta}}{k}= l_1\circ\dots\circ l_s$, where now each layer is a function $l_i:[\mathbb{Z}]_k^{n_i}\rightarrow [\mathbb{Z}]_k^{n_{i+1}}$ that operates over $k$-bit bitvectors and is defined as follows:
\begin{align}
    x'_i &= \sum_{j=1}^{n_i} w_{ij} x_j + b_i,\label{eq:sum}\\
    x''_i &= \text{round}(x'_i,M_i) = \lfloor x'_i\cdot M_i \rfloor,  \qquad \text{and} \label{eq:round} \\
    y_i &= \sigma_i(\min\{2^{N_i}-1, x''_i \}),\label{eq:relun}
\end{align}
Here, $w_{i,j}\in [\mathbb{Z}]_k^{n_i}$ and $b_i\in [\mathbb{Z}]_k^{n_i}$ for each $1\leq j\leq n_i$ and $1\leq i\leq n_0$ denote the weights and biases of $f$ which are also bitvectors of appropriate dimension. Note that it is a task of the training procedure to ensure that trained weights and biases are bitvectors, see below. 
In eq.~\eqref{eq:sum}, the linear map defined by weights $w_{i,j}$ and biases $b_i$ is applied to the input values $x_j$. Then, eq.~\eqref{eq:round} multiplies the result of eq.~\eqref{eq:sum} by $M_i$ and takes the floor of the obtained result. This is done in order to scale the result and round it to the nearest valid fixed-point value, for which one typically uses $M_i$ of the form $2^{-k}$ for some integer $k$. Finally, eq.~\eqref{eq:relun} applies an activation function $\sigma_i$ to the result of eq.~\eqref{eq:round} where the result is first ``cut-off'' if it exceeds $2^{N_i}-1$, i.e., to avoid integer overflows, and then passed to the activation function. We restrict ourselves to {\em monotone} activation functions, which will be necessary for our IBP procedure to be correct. This is still a very general assumption which includes a rich class of activation function, e.g.~ReLU, sigmoid or tanh activation functions. Furthermore, similarly to most quantization-aware training procedures our method assumes that it is provided with {\em quantized} versions of these activation functions that operate over bit-vectors.


\paragraph{Adversarial robustness for QNNs} We now formalize the notion of adversarial robustness for QNN classifiers. Let $\semink{f_{\theta}}{k}: [\mathbb{Z}]_k^{n}\rightarrow [\mathbb{Z}]_k^{m}$ be a $k$-bit QNN.
It naturally defines a classifier with $m$ classes by assuming that it assigns to an input $x\in [\mathbb{Z}]_k^{n}$ a label of the maximal output neuron on input value $x$, i.e.~$y = \mathsf{class}(x) = \mathsf{argmax}_{1\leq i\leq m} \semink{f_{\theta}}{k}(x)[i]$ with $\semink{f_{\theta}}{k}(x)[i]$ being the value of the $i$-th output neuron on input value $x$. If the maximum is attained at multiple output neurons, we assume that $\mathsf{argmax}$ picks the smallest index $1\leq i\leq m$ for which the maximum is attained.

Intuitively, a QNN is adversarially robust at a point $x$ if it assigns the same class to every point in some neighbourhood of $x$. Formally, given $\epsilon>0$, we say that $\semink{f_{\theta}}{k}$ is {\em $\epsilon$-adversarially robust} at point $x$ if
\[ \forall x'\in [\mathbb{Z}]_k^{n}.\, ||x-x'||_\infty < \epsilon \Rightarrow \mathsf{class}(x') = \mathsf{class}(x), \]
where $||\cdot||_\infty$ denotes the $L_\infty$-norm. Then, given a finite dataset $\mathcal{D} = \{(x_1,y_1),\dots,(x_{|\mathcal{D}|},y_{|\mathcal{D}|})\}$ with $x_i\in[\mathbb{Z}]_k^{n}$ and $y_i\in[\mathbb{Z}]_k^{m}$ for each $1\leq i\leq |\mathcal{D}|$, we say that $\semink{f_{\theta}}{k}$ is {\em $\epsilon$-adversarially robust} with respect to the dataset $\mathcal{D}$ if it is $\epsilon$-adversarially robust at each datapoint in $\mathcal{D}$.


\section{Quantization-aware Interval Bound Propagation}\label{sec:training}

In this section, we introduce an end-to-end differentiable abstract interpretation method for training certifiably robust QNNs. We achieve this by extending the interval bound propagation (IBP) method of~\cite{gowal2019scalable} to the discrete semantic of QNNs.
Our resulting quantization-aware interval bound propagation method (QA-IBP) trains an interval arithmetic abstraction of a QNN via stochastic gradient descent by propagating upper and lower bounds for each layer instead of concrete values.

First, we replace each layer $l_i$ with two functions $\underline{l_i},\, \overline{l_i}$: $[\mathbb{Z}]_k^{n_i} \rightarrow [\mathbb{Z}]_k^{n_0}$ defined as follows:
\begin{align}
    \mu_j &= \frac{\overline{x_j} + \underline{x_j}}{2} & r_j &= \frac{\overline{x_j} - \underline{x_j}}{2}\\
    \mu_i &= \sum_{j=1}^{n_i} w_{ij} \mu_j + b_i & r_i &= \sum_{j=1}^{n_i} | w_{ij} |r_j  \\
    \underline{x'_i} &= \mu_i - r_i & \overline{x'_i} &= \mu_i + r_i \label{eq:ibpsum}\\
    \underline{x''_i} &= \text{round}(\underline{x'_i},k_i) = \lfloor \underline{x'_i}\cdot M_i \rfloor \\
    \overline{x''_i} &= \text{round}(\overline{x'_i},k_i) = \lfloor \overline{x'_i}\cdot M_i \rfloor \label{eq:ibpround}\\
    \underline{y_i} &= \max\{0,\min\{2^{N_i}-1, \underline{x''_i} \}\} \\
    \overline{y_i} &= \max\{0,\min\{2^{N_i}-1, \overline{x''_i} \}\}.
\end{align}
As with standard QNNs, $w_{i,j}\in [\mathbb{Z}]_k^{n_i}$ and $b_i\in [\mathbb{Z}]_k^{n_i}$ for each $1\leq j\leq n_i$ and $1\leq i\leq n_0$ denote the weights and biases of $f$ which are also bit-vectors of appropriate dimension. 
By the sequential composition of all layers of the QNN we get the IBP representation of the QNN in the form of two functions $\underline{\semink{f_{\theta}}{k}}$ and $\overline{\semink{f_{\theta}}{k}}$. For a given input sample $x$ and $\epsilon>0$ we can use the IBP representation of $\semink{f_{\theta}}{k}$ to potentially prove the adversarial robustness of the QNN. In particular, the input sample defines an abstract interval domain $(\underline{x},\overline{x})$ with $\underline{x} = x-\epsilon$ and $\overline{x} = x + \epsilon$.
Next, we propagate the abstract domains through the IBP representation of the network to obtain output bounds $\underline{y} =\underline{\semink{f_{\theta}}{k}}(\underline{x},\overline{x})$ and $\overline{y} =\overline{\semink{f_{\theta}}{k}}(\underline{x},\overline{x})$.
Finally, we know that $\semink{f_{\theta}}{k}$ is {\em $\epsilon$-adversarially robust} at point $x$ if
\begin{equation}\label{eq:ibprobust}
    \underline{y_i} > \overline{y_j}, \qquad\text{for } i=\mathsf{class}(x) \text{ and } \forall j\neq \mathsf{class}(x).
\end{equation}

\begin{figure}
    \centering
    \includegraphics[width=0.45\textwidth]{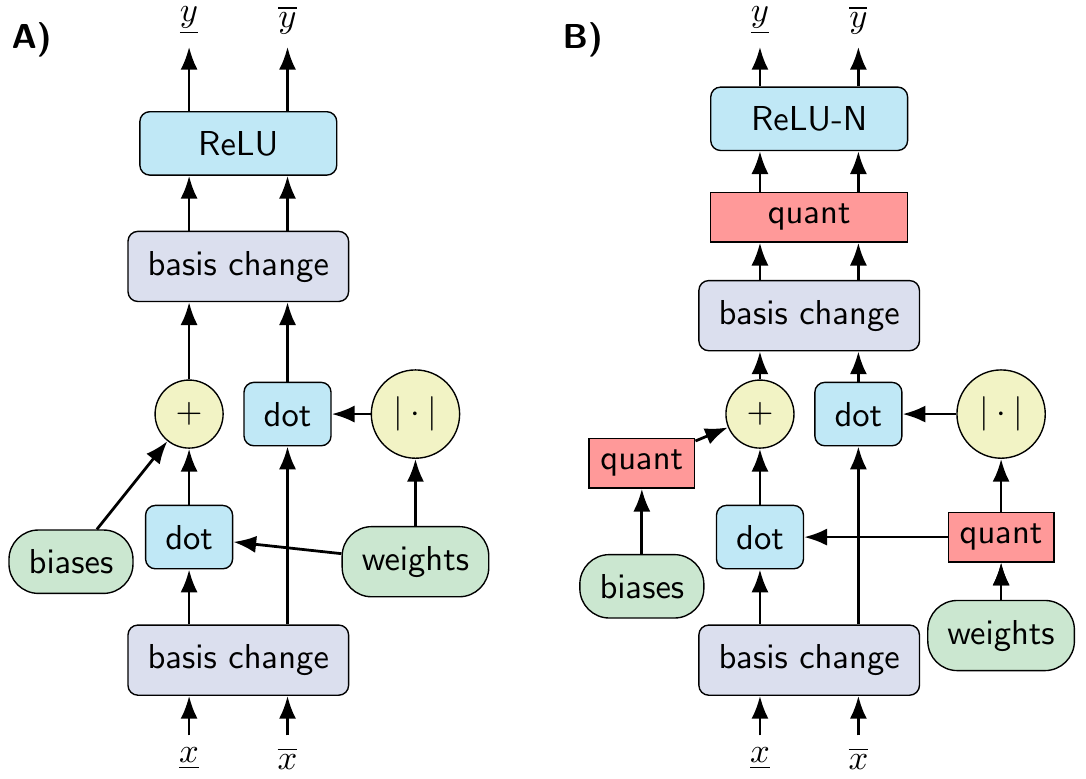}
    \caption{Illustration of how the interval bound propagation training graph is affected by quantization-aware training. \textbf{A)} Standard IBP inference graph. \textbf{B)} IBP inference path with fake quantization operations inserted to model quantized weights, biases, and computations.}
    \label{fig:fakequant}
\end{figure}

\paragraph{End-to-end differentiation}
We modify the IBP representation of QNNs described above to allow an end-to-end differentiation necessary for a stochastic gradient descent-based learning algorithm. In particular, first, we apply the straight-through gradient estimator to propagate the error backward through the rounding operations in Eq. \ref{eq:ibpround}. We do this by replacing the non-differentiable rounding operation with the fake quantization function 
\begin{align}
    \text{fake\_quant}(x_i) &:= \text{round}(x_i,k_i)\\
    \frac{\partial\ \text{fake\_quant}(x_i)}{\partial x_i} &:= 1.
\end{align}
We also add fake quantization operations around the weights and biases in Eq. \ref{eq:ibpsum}. The modified training graph is visualized in Figure \ref{fig:fakequant}.

For a single training sample $(x,j)$, we define the per-sample training loss
\begin{align}
    L(\underline{y},\overline{y},j) = \sum_{i\neq j}(\overline{y_i} - \underline{y_j})\Ind[\underline{y_j}-\overline{y_i}\leq 0],
\end{align}
where $\underline{y}, \overline{y}$ are the QA-IBP output bounds QNN with respect to the input domain $(x-\epsilon,x+\epsilon)$ and $j$ corresponds to the label, i.e., class $j$.
The loss term encourages the QA-IBP to produce output bounds that prove adversarial robustness of the QNN $\semink{f_{\theta}}{k}$ with respect to the input $x$ and adversarial radius $\epsilon$.

\paragraph{Existence of robust QNNs} We conclude this section by presenting an interesting result on the existence of robust QNNs. In particular, given $\epsilon>0$ and a finite dataset of $1$-dimensional bit-vectors whose any two distinct datapoints are at least $2\epsilon$ away, we prove that there exists a QNN with ReLU activations that is $\epsilon$-robust with respect to the dataset. Furthermore, we provide an upper bound on the number of neurons that the QNN must contain. We assume $2\epsilon$ distance simply for the $\epsilon$-neighbourhoods of datapoints to be disjoint so that robustness cannot impose contradicting classification conditions.

Note that this result {\em does not} hold for real arithmetic feed-forward neural networks with ReLU or any other continuous activation functions. Indeed, it was observed in~\cite[Corollary~5.12]{MirmanBV21} that a dataset $\{(-2,-1), (0,1), (2,-1)\}$ cannot be $1$-robustly classified by a feed-forward neural network which uses continuous activation functions. The intuition behind this impossibility result for real arithmetic networks is that a classifier would have to be a continuous function that correctly classifies all points in some open neighbourhoods of $x=-1$ and $x=1$.

\begin{theorem}\label{thm:existence}
Let $\epsilon>0$ and let $\mathcal{D} = \{(x_1,y_1),\dots,(x_{|\mathcal{D}|},y_{|\mathcal{D}|})\}$ with $x_i\in[\mathbb{Z}]_k^1$ and $y_i\in[\mathbb{Z}]_k^1$ for each $1\leq i\leq |\mathcal{D}|$. Suppose that $||x_i - x_j||_{\infty}\geq 2\epsilon$ for each $i\neq j$. Then, there exists a QNN $\semink{f_{\theta}}{k}$ which is $\epsilon$-robust with respect to the dataset $\mathcal{D}$ and which consists of $\mathcal{O}(|\mathcal{D}| \cdot \lceil\epsilon\cdot 2^k + 1 \rceil)$ neurons.
\end{theorem}

The proof of Theorem~\ref{thm:existence} is provided in the Appendix. It starts with an observation that the set $\mathcal{B}_{\epsilon}(x_i)=\{x'\in[\mathbb{Z}]_k^1 \mid ||x'-x_i||_{\infty}<\epsilon\}$ is finite and consists of at most $\lceil 2\epsilon\cdot 2^k+1 \rceil$ bit-vectors for each $1\leq i\leq |\mathcal{D}|$. Hence, constructing a QNN $\semink{f_{\theta}}{k}$ that is $\epsilon$-robust with respect to the dataset $\mathcal{D}$ is equivalent to constructing a QNN that correctly classifies $\mathcal{D}' = \cup_{i=1}^{|\mathcal{D}|} \{(x',y_i) \mid x'\in \mathcal{B}_{\epsilon}(x_i)\}$
which consists of $\mathcal{O}(|\mathcal{D}| \cdot \lceil\epsilon\cdot 2^k + 1 \rceil)$ datapoints. In the Appendix, we then design the QNN $\semink{f_{\theta}}{k}$ that correctly classifies a dataset and consists of at most linearly many neurons in the dataset size. Our construction also implies the following corollary.

\begin{corollary}
Let $\mathcal{D} = \{(x_1,y_1),\dots,(x_{|\mathcal{D}|},y_{|\mathcal{D}|})\}$ with $x_i\in[\mathbb{Z}]_k^1$ and $y_i\in[\mathbb{Z}]_k^1$ for each $1\leq i\leq |\mathcal{D}|$. Then, there exists a QNN $\semink{f_{\theta}}{k}$ which correctly classifies the dataset, i.e.~$\semink{f_{\theta}}{k}(x_i) = y_i$ for each $1\leq i\leq |\mathcal{D}|$, and which consists of $\mathcal{O}(|\mathcal{D}|)$ neurons.
\end{corollary}


\section{A Complete Decision Procedure for QNN Verification}\label{sec:verification}

In the previous section, we presented a quantization-aware training procedure for QNNs with robustness guarantees which was achieved by extending IBP to quantized neural network interpretations. We now show that IBP can also be used towards designing a {\em complete} verification procedure for already trained feed-forward QNNs. By completeness, we mean that the procedure is guaranteed to return either that the QNN is robust or to produce an adversarial attack.

There are two important novel aspects of the verification procedure that we present in this section. First, to the best of our knowledge this is the first complete robustness verification procedure for QNNs that is applicable to networks with non-piecewise linear activation functions. Existing constraint solving based methods that reduce verification to SMT-solving are complete but they only support piecewise linear activation fucntions such as ReLU~\citep{krizhevsky2010convolutional}. These could in theory be extended to more general activation functions by considering more expressive satisfiability modulo theories~\citep{smt2018handbook}, however this would lead to inefficient verification procedures and our experimental results in the following section already demonstrate the significant gain in scalability of our IBP-based methods as opposed to SMT-solving based methods for ReLU networks. Second, we note that while our IBP-based verification procedure is complete for QNNs, in general it is known that existing IBP-based verification procedures for real arithmetic neural networks are {\em not complete}~\citep{mirman2021fundamental}. Thus, our result leads to an interesting contrast in IBP-based robustness verification procedures for QNNs and for real arithmetic neural networks.

\paragraph{Verification procedure} We now describe our robustness verification procedure for QNNs. Its pseudocode is shown in Algorithm~\ref{alg:algorithm}. Since verifying $\epsilon$-robustness of a QNN with respect to some finite dataset $\mathcal{D}$ and $\epsilon>0$ is equivalent to verifying $\epsilon$-robustness of the QNN at each datapoint in $\mathcal{D}$, Algorithm~\ref{alg:algorithm} only takes as inputs a QNN $\semink{f_{\theta}}{k}$ that operates over bit-vectors of bit-width $k$, a single datapoint $x\in [\mathbb{Z}]_k^{n}$ and a robustness radius $\epsilon>0$. It then returns either \texttt{ROBUST} if $\semink{f_{\theta}}{k}$ is verified to be $\epsilon$-robust at $x\in [\mathbb{Z}]_k^{n}$, or \texttt{VULNERABLE} if an adversarial attack $||x'-x||_{\infty} < \epsilon$ with $\mathsf{class}(x') = \mathsf{class}(x)$ is found.

The algorithm proceeds by initializing a stack $D$ of abstract intervals to contain a single element $\{(x-\epsilon\cdot \mathbf{1}, x+\epsilon \cdot \mathbf{1})\}$, where $\mathbf{1}\in [\mathbb{Z}]_k^{n}$ is a unit bit-vector of bit-width $k$. Intuitively, $D$ contains all abstract intervals that may contain concrete adversarial examples but have not yet been processed by the algorithm. The algorithm then iterates through a loop which in each loop iteration processes the top element of the stack. Once the stack is empty and the last loop iteration terminates, Algorithm~\ref{alg:algorithm} returns \texttt{ROBUST}.

In each loop iteration, Algorithm~\ref{alg:algorithm} pops an abstract interval $(\underline{x},\overline{x})$ from $D$ and processes it as follows. First, it uses IBP for QNNs that we introduced before to propagate $(\underline{x},\overline{x})$ in order to compute an abstract interval $(\underline{y},\overline{y})$ that overapproximates the set of all possible outputs for a concrete input point in $(\underline{x},\overline{x})$. The algorithm then considers three cases. First, if $(\underline{y},\overline{y})$ does not violate Equation~\eqref{eq:ibprobust} which characterizes violation of robustness by a propagated abstract interval, Algorithm~\ref{alg:algorithm} concludes that the abstract interval $(\underline{x},\overline{x})$ does not contain an adversarial example and it proceeds to processing the next element of $D$. Second, if $(\underline{y},\overline{y})$ violates Equation~\eqref{eq:ibprobust}, the algorithm uses projected gradient descent restricted to $(\underline{x},\overline{x})$ to search for an adversarial example and returns \texttt{VULNERABLE} if found. Note that the adversarial attack is generated with respect to the quantization-aware representation of the network, thus ensuring that the input space corresponds to valid quantized inputs. Third, if $(\underline{y},\overline{y})$ violates Equation~\eqref{eq:ibprobust} but the adversarial attack could not be found by projected gradient descent, the algorithm refines the abstract interval $(\underline{x},\overline{x})$ by splitting it into two smaller subintervals. This is done by identifying
\[ i^{\ast} = \mathsf{argmax}_{1\leq i\leq n} (\overline{x}[i] - \underline{x}[i]) \]
and splitting the abstract interval $(\underline{x},\overline{x})$ along the $i^{\ast}$-th dimension into two abstract subintervals $(\underline{x'},\overline{x'}),(\underline{x''},\overline{x''})$, which are both added to the stack $D$.

\paragraph{Correctness, termination and completeness} The following theorem establishes that Algorithm~\ref{alg:algorithm} is complete, that it terminates on every input and that it is a complete robustness verification procedure. The proof is provided in the Appendix.

\begin{theorem}\label{thm:verification}
If Algorithm~\ref{alg:algorithm} returns \texttt{ROBUST} then $\semink{f_{\theta}}{k}$ is $\epsilon$-robust at $x\in [\mathbb{Z}]_k^{n}$. On the other hand, if Algorithm~\ref{alg:algorithm} returns \texttt{VULNERABLE} then there exists an adversarial attack $||x'-x||_{\infty} < \epsilon$ with $\mathsf{class}(x') \neq \mathsf{class}(x)$. Therefore, Algorithm~\ref{alg:algorithm} is correct. Furthermore, Algorithm~\ref{alg:algorithm} terminates and is guaranteed to return an output on any input. Since Algorithm~\ref{alg:algorithm} is correct and it terminates, we conclude that it is also complete.
\end{theorem}

\begin{algorithm}[t]
\caption{QNN robustness verification procedure}
\label{alg:algorithm}
\begin{algorithmic}[1]
\State \textbf{Input} QNN $\semink{f_{\theta}}{k}$, datapoint $x\in [\mathbb{Z}]_k^{n}$, robustness radius $\epsilon>0$
\State \textbf{Output} \texttt{ROBUST} or \texttt{VULNERABLE}
\State $D \leftarrow \{(x-\epsilon\cdot \mathbf{1},x+\epsilon\cdot \mathbf{1})\}$
\While{$D \neq \{\}$}
\State $(\underline{x},\overline{x}) \leftarrow$ pop item from $D$
\State $(\underline{y},\overline{y}) \leftarrow  \underline{\semink{f_{\theta}}{k}}(\underline{x},\overline{x}), \overline{\semink{f_{\theta}}{k}}(\underline{x},\overline{x})$ propagated via IBP
\If{$(\underline{y},\overline{y})$ violates Equation (\ref{eq:ibprobust})}
\State Try to generate adversarial example using projected gradient descent restricted to $(\underline{x},\overline{x})$
\If{adversarial example found}
\State \textbf{return} \texttt{VULNERABLE}
\Else
\State $(\underline{x'},\overline{x'}),(\underline{x''},\overline{x''}) \leftarrow$ partition of $(\underline{x},\overline{x})$ into two abstract subintervals
\State $D \leftarrow D \cup \{(\underline{x'},\overline{x'}),(\underline{x''},\overline{x''})\}$
\EndIf
\EndIf
\EndWhile
\State \textbf{return} \texttt{ROBUST}
\end{algorithmic}
\end{algorithm}

\section{Experiments}\label{sec:exp}
We perform an experimental evaluation to assess the effectiveness of our quantization-aware interval bound propagation (QA-IBP). 
In particular, we first use our training procedure to train two QNNs. We then use our complete verification procedure to verify robustness of trained QNNs and we compare our method to the existing verification methods for QNNs of \cite{giacobbe2020many} and \cite{henzinger2021scalable}. 
Our full experimental setup and code can be found on GitHub \footnote{\url{https://github.com/mlech26l/quantization_aware_ibp}}.

\begin{table*}
    \centering
    \begin{tabular}{l|ccc|ccc}\toprule
        Method &  \multicolumn{3}{c}{MNIST} & \multicolumn{3}{c}{Fashion-MNIST} \\
        & $\epsilon=0$ & $\epsilon=1$ & $\epsilon=4$ & $\epsilon=0$ & $\epsilon=1$ & $\epsilon=4$ \\\midrule
        QF\_BV SMT \cite{giacobbe2020many} & 97.1\% & 92\% & 0\% & 85.6\% & 44\% & 0\% \\
        QF\_BV SMT \cite{henzinger2021scalable} &  97.1\% & \textbf{99\%}$^*$ & 53\%  & 85.6\% & 76\% & 36\% \\
        QA-IBP (ours) & 99.2\% & \textbf{98.8\%} & \textbf{95.6\%} & 86.3\% & \textbf{80.0\%} & \textbf{59.8\%}\\\bottomrule
    \end{tabular}
    \caption{Certified robust accuracy of a convolutional neural network trained with QA-IBP compared to existing methods for certifying $L_\infty$-robustness of quantized neural networks reported in the literature. $^*$ Note that \cite{henzinger2021scalable} certified only a subset of the test set due to a high per-sample runtime of their approach. Due to this choice they reported a higher $\epsilon=1$ robust accuracy than clean accuracy.}
    \label{tab:experiment}
\end{table*}

We train two CNNs with QA-IBP using an 8-bit quantization scheme for both weights and activations on the MNIST \cite{lecun1998gradient} and Fashion-MNIST \cite{xiao2017online} datasets. We quantize all layers; however, we note that our approach is also compatible with mixing quantized and non-quantized layers. Our MNIST network consists of five convolutional layers, followed by two fully-connected layers. Our Fashion-MNIST network contains three convolutional layers and two fully-connected layers. We apply a fixed pre-defined fixed-point format on both weights and activations.  Further details on the network architectures can be found in the Appendix.
We use the Adam optimizer \cite{kingma2014adam} with a learning rate of $10^{-4}$ with decoupled weight decay of $10^{-4}$ \cite{loshchilov2018decoupled} and a batch size of 512. We train our networks for a total of 5,000,000 gradient steps on a single GPU.
We linearly scale the value of $\epsilon$ during the QA-IBP training from 0 to 4 and apply the elision of the last layer optimization as reported in \cite{gowal2019scalable}. 
We pre-train our networks for 5000 steps in their non-IBP representation with a learning rate of $5\cdot 10^{-4}$.

After training, we certify all test samples using Algorithm \ref{alg:algorithm} with a timeout of 20s per sample. We certify for $L_\infty$-robustness with radii 1 and 4 to match the experimental setup of \cite{henzinger2021scalable}.
We report and compare the certified robust accuracy of the trained networks to the values reported in \cite{henzinger2021scalable}.

The results in Table \ref{tab:experiment} demonstrate that our QA-IBP significantly improves the certified robust accuracy for QNNs on both datasets.

Regarding the runtime of the verification step, both \cite{giacobbe2020many} and \cite{henzinger2021scalable} certified a smaller model compared to our evaluation. In particular, \cite{giacobbe2020many} report a mean runtime of their 8-bit model to be over 3 hours, and \cite{henzinger2021scalable} report the mean runtime of their 6-bit model to be 90 and 49 seconds for MNIST/Fashion-MNIST, respectively.
Conversely, our method was evaluated with a timeout of 20 seconds and tested on larger networks than the two existing approaches, showing the efficiency of our approach.

\subsection{Ablation analysis}
In this section, we assess the effectiveness of Algorithm \ref{alg:algorithm} for certifying robustness compared to an incomplete verification based on the QA-IBP output bounds alone.
In particular, we perform an ablation and compare Algorithm \ref{alg:algorithm} to an incomplete verification baseline. Our baseline consists of checking the bounds obtained by QA-IBP as an incomplete verifier combined with projected gradient descent (PGD) as an incomplete falsifier, i.e., we try to certify robustness via QA-IBP and simultaneously try to generate an adversarial attack using PGD.
In particular, this baseline resembles a version of Algorithm \ref{alg:algorithm} without any branching into subdomains.
We carry our ablation analysis on the two networks of our first experiment. We report the number of samples where the verification approach could not determine whether the network is robust on the sample or not. We use a timeout of 20s per instance when running our algorithm. 
\begin{table}[]
    \centering
    \begin{tabular}{c|ccc}\toprule
        & \multicolumn{3}{c}{CIFAR-10} \\
        Weight decay & $\epsilon=0$ & $\epsilon=1$ & $\epsilon=4$  \\\midrule
        $1\cdot 10^{-4}$ & 30.0\% & 28.4\% & \textbf{22.4\%} \\
        $5\cdot 10^{-5}$ & 39.6\% & \textbf{34.1\%} & 20.6\% \\
        $1\cdot 10^{-5}$ & \textbf{83.1\%} & 0\% & 0\% \\\bottomrule
    \end{tabular}
    \caption{Robust accuracy of our convolutional neural network trained with QA-IBP on the CIFAR-10 dataset with various values for the weight decay. The results express the \textit{robustness-accuracy tradeoff} \cite{tsipras2018robustness}, i.e., the empirical observation made for non-quantized neural networks that we can have a high robustness or a high accuracy but not both at the same time.}
    \label{tab:cifar10}
\end{table}

The results shown in Table \ref{tab:timeout} indicate that our algorithm is indeed improving on the number of samples for which robustness could be decided. 
However, the improvement is relatively small, suggesting that the QA-IBP training stems from most of the observed gains.
\begin{table*}[]
    \centering
    \begin{tabular}{l|cc|cc}\toprule
         Method & \multicolumn{2}{c}{MNIST} & \multicolumn{2}{c}{Fashion-MNIST}  \\
         & $\epsilon=1$ & $\epsilon=4$ & $\epsilon=1$ & $\epsilon=4$ \\\midrule
      IBP + Projected gradient descent &  \textbf{0.35\%} & 3.51\% & 3.96\% & 19.23\%  \\ 
      Algorithm \ref{alg:algorithm} & \textbf{0.35\%} & \textbf{3.47\%} & \textbf{3.93\%} & \textbf{18.51\%} \\\bottomrule
    \end{tabular}
    \caption{Percentage of samples where the robustness of the network can be determined, i.e., certified or falsified, by the method (lower is better). Algorithm timeout was set to 20s per instance. }
    \label{tab:timeout}
\end{table*}

\subsection{Limitations}
In this section, we aim to scale our QA-IBP beyond the two gray-scale image classification tasks studied before to the CIFAR-10 dataset \cite{krizhevsky2010convolutional}.
Our setup consists of the same convolutional neural network as used for the Fasion-MNIST.
We run our setup with several values for the weight decay rate. Similar to above, we linearly scale $\epsilon$ from 0 to 4 during training and report the certified robust accuracy obtained by QA-IBP with an $\epsilon$ of 1 and 4 and the clean accuracy.

The results in Table \ref{tab:cifar10} express the robustness-accuracy tradeoff, i.e., the observed antagonistic relation between clean accuracy and robustness, which has been extensively studied in non-quantized neural networks \cite{zhang2019theoretically,bubeck2021universal}.
Depending on the weight decay value, different points on the trade-off were observed. In particular, our training procedure either obtains a network that has an acceptable accuracy but no robustness or a certifiable robust network with a significantly reduced clean accuracy. 

We also trained larger models on all datasets (MNIST, Fashion-MNIST, and CIFAR-10), but observed the same behavior of having a high accuracy but no robustness when training, i.e., as in row with $1\cdot 10^{-5}$ in Table \ref{tab:cifar10}.

We found the underlying reason for this behavior to be activations of internal neurons that clamp to the minimum and maximum value of the quantization range in their QA-IBP representation but not in their standard representation. Consequently, the activation gradients become zero during QA-IBP due to falling in the constant region of the activation function. 
This effect is specific to quantized neural networks due to the upper bound on the representative range of values.
Our observation hints that future research needs to look into developing dynamic quantization ranges or weight decay schedules that can adapt to both the standard and the QA-IBP representation of a QNN.

\section{Conclusion}
In this paper, we introduced quantization-aware interval bound propagation (QA-IBP), the first method for training certifiably robust QNNs. We also present a theoretical result on the existence and upper bounds on the needed size of a robust QNN for a given dataset of $1$-dimensional datapoints. Moreover, based on our interval bound propagation method, we developed the first complete verification algorithm for QNNs that may be run on GPUs. 
We experimentally showed that our training scheme and verification procedure advance the state-of-the-art on certifying $L_\infty$-robustness of QNNs. 

We also demonstrated the limitations of our method regarding training stability and convergence. Particularly, we found that the boundedness of the representable value range of QNNs compared to standard networks leads to truncation of the abstract domains, which in turn leads to gradients becoming zero. Our results suggest that dynamic quantization schemes that adapt their quantization range to the abstract domains instead of the concrete activation values of existing quantization schemes may further improve the certified robust accuracy of quantized neural networks. 

Nonetheless, our work serves as a new baseline for future research. Promising directions on how to improve upon QA-IBP and potentially overcome its numerical challenges is to adopt advanced quantization-aware training techniques. For instance, dynamical quantization ranges \cite{choi2018pact, jacob2018quantization},  mixed-precision layers \cite{zhou2016dorefa,yao2021hawq}, and automated architecture search \cite{tan2019efficientnet} have shown promising results for standard training QNNs and might enhance QA-IBP-based training procedures as well.
Moreover, further improvements may be feasible by adapting recent advances in IBP-based training methods for non-quantized neural networks \cite{muller2022certified} to our quantized IBP variant.

\FloatBarrier
\appendix

 \bibliography{aaai23}

\section{Acknowledgments}
This work was supported in part by the ERC-2020-AdG 101020093, ERC CoG 863818 (FoRM-SMArt) and the European Union’s Horizon 2020 research and innovation programme under the Marie Skłodowska-Curie Grant Agreement No.~665385.
Research was sponsored by the United States Air Force Research Laboratory and the United States Air Force Artificial Intelligence Accelerator and was accomplished under Cooperative Agreement Number FA8750-19-2-1000. The views and conclusions contained in this document are those of the authors and should not be interpreted as representing the official policies, either expressed or implied, of the United States Air Force or the U.S. Government. The U.S. Government is authorized to reproduce and distribute reprints for Government purposes notwithstanding any copyright notation herein. The research was also funded in part by the AI2050 program at Schmidt Futures (Grant G-22-63172) and Capgemini SE.

\section{Appendix}

\subsection{Proof of Theorem~1}\label{app:proofexistence}

Note that, for each $1\leq i\leq |\mathcal{D}|$, the set $\mathcal{B}_{\epsilon}(x_i)=\{x'\in[\mathbb{Z}]_k^1 \mid ||x'-x_i||_{\infty}<\epsilon\}$ is finite and consists of at most $\lceil 2\epsilon\cdot 2^k+1 \rceil$ bit-vectors. Hence, constructing a QNN $\semink{f_{\theta}}{k}$ that is $\epsilon$-robust with respect to the dataset $\mathcal{D}$ is equivalent to constructing a QNN that correctly classifies the dataset
\[ \mathcal{D}' = \bigcup_{i=1}^{|\mathcal{D}|} \{(x',y_i) \mid x'\in \mathcal{B}_{\epsilon}(x_i)\} \]
that consists of $\mathcal{O}(|\mathcal{D}| \cdot \lceil\epsilon\cdot 2^k + 1 \rceil)$ datapoints. Let $N = |\mathcal{D}'|$. In what follows, we fix an enumeration $\{(x'_1, y'_1), \dots, (x'_N, y'_N)\}$ of datapoints in $\mathcal{D}'$.

We design the QNN $\semink{f_{\theta}}{k}$ so that it consists of the input layer with $1$ neuron, $1$ hidden layer with $5N$ neurons and the output layer with $1$ neurons. The neurons in the hidden layer are partitioned into $ N$ gadgets $G_i$ with $1\leq i\leq N$, where each gadget consists of $5$ neurons. For each $G_i$, we connect the input layer neuron to the $5$ neurons in the gadget and we connect the $5$ neurons in the gadget to output layer neuron. We then use the connecting edges in order to add the following expression to value of the output layer neuron, whenever the value of the input layer neuron is equal to $z$:
\begin{equation}\label{eq:proof}
    y'_i \cdot \Big(\mathsf{ReLU}(z - x'_i + 1) - \mathsf{ReLU}(z - x'_i) + \mathsf{ReLU}(x'_i - z) - \mathsf{ReLU}(x'_i - z - 1) - 1\Big).
\end{equation}
One can verify by inspection that the expression in eq.~\eqref{eq:proof} evaluates to $y'_i$ if $z = x'_i$ and to $0$ otherwise and that it can indeed be encoded by $5$ neurons in the hidden layer. Hence, if we consider the output layer neuron and take the weighted sum of all incoming edges from each gadget, this construction ensures that for any input neuron value $z \in [\mathbb{Z}]_k^{n}$ we have that the value of the output neuron of $\semink{f_{\theta}}{k}$ is equal to
\[ \semink{f_{\theta}}{k}(z) = \sum_{i=1}^N y'_i \cdot \mathbb{I}(z = x'_i) \]
with $\mathbb{I}$ an indicator function. Therefore, as datapoints in $\mathcal{D}'$ are distinct, one may conclude that $\semink{f_{\theta}}{k}(x'_i) = y'_i$ for each $1\leq i\leq N$, as desired.

\subsection{Proof of Theorem~2}\label{app:proofcompleteness}

We first show that Algorithm~1 is correct, i.e.~that if it returns \texttt{ROBUST} then $\semink{f_{\theta}}{k}$ is $\epsilon$-robust at $x\in [\mathbb{Z}]_k^{n}$ and that if it returns \texttt{VULNERABLE} then there exists an adversarial attack $||x'-x||_{\infty} < \epsilon$ with $\mathsf{class}(x') = \mathsf{class}(x)$. Indeed, the correctness of \texttt{ROBUST} outputs follows from the fact that, in order for Algorithm~1 to terminate, it had to split the initial abstract interval $\{(x-\epsilon\cdot \mathbf{1}, x+\epsilon \cdot \mathbf{1})\}$ into a finite number of abstract subintervals and to show that their propagations computed via IBP do not contain outputs of a different class. Therefore, the correctness of \texttt{ROBUST} follows by the correctness of our IBP for QNNs in Section~\ref{sec:training}. The fact that \texttt{VULNERABLE} outputs are correct follows from the fact that computed adversarial examples can be easily verified by feeding them to $\semink{f_{\theta}}{k}$ and by checking that the class of the computed output differs from $\mathsf{class}(x)$.

To show that Algorithm~1 terminates, observe that in every loop iteration Algorithm~1 either pops an item from $D$ and does not add new items, it returns \texttt{VULNERABLE} or it pops an item from $D$ and replaces it by two new abstract intervals that are obtained by splitting the popped one. Hence, all three cases preserve the invariant that $\sum_{(\underline{x},\overline{x})\in D}\#((\underline{x},\overline{x}))^2$ strictly decreases between any two consecutive iterations, where we use $\#((\underline{x},\overline{x}))$ to denote the number of bit-vectors contained in $(\underline{x},\overline{x})$. Hence, as the number of bit-vectors of bit-width $k$ that are contained in the initial abstract interval $\{(x-\epsilon\cdot \mathbf{1},x+\epsilon\cdot \mathbf{1})\}$ is finite, we conclude that Algorithm~1 must terminate and return an output in at most finitely many steps.

Finally, since Algorithm~1 terminates it must return either an output \texttt{ROBUST} or \texttt{VULNERABLE} on any input. On the other hand, since Algorithm~1 is complete, its returned output is always correct. Hence, Algorithm~1 is guaranteed to return either that the QNN is robust or to prove the existence an adversarial attack, and therefore it is complete. This concludes the proof of Theorem~2.

\subsection{Experimental details}\label{app:experiment}
The network architectured used in our experiments can be found in Table \ref{tab:archmnist}, \ref{tab:archfashion}, and \ref{tab:archcifar}.
For the MNIST network, we used a Q3.5 fixed-point format for the activations, Q5.3 for the bias terms, and Q2.6 format for the weights. 
For both the Fashion-MNIST and the CIFAR-10 networks, we used a Q4.4 fixed-point format for the activations, Q5.3 for the bias terms, and Q2.6 format for the weights. 
\begin{table}[]
    \centering
    \begin{tabular}{c|c}\toprule
Layer & Parameters \\\midrule
     Conv2D & F=64, K=5, S=2, ReLU-N \\
     Conv2D & F=128, K=3, S=1, ReLU-N\\ 
     Conv2D & F=256, K=3, S=1, ReLU-N\\ 
     Conv2D & F=384, K=3, S=1, ReLU-N\\ 
     Conv2D & F=512, K=3, S=2, ReLU-N\\
     Flatten & \\
     Fully-connected & U=128, ReLU-N\\
     Fully-connected & U=10\\\bottomrule
\end{tabular}
    \caption{Convolutional neural network architecture used for the MNIST experiment. F represents the number of filters, K the kernel size, S the stride, and U the number of units.}
    \label{tab:archmnist}
\end{table}
\begin{table}[]
    \centering
    \begin{tabular}{c|c}\toprule
Layer & Parameters \\\midrule
     Conv2D & F=64, K=5, S=2, ReLU-N \\
     Conv2D & F=96, K=3, S=1, ReLU-N\\ 
     Conv2D & F=128, K=3, S=2, ReLU-N\\ 
     Flatten & \\
     Fully-connected & U=128, ReLU-N\\
     Fully-connected & U=10\\\bottomrule
\end{tabular}
    \caption{Convolutional neural network architecture used for the Fashion-MNIST experiment. F represents the number of filters, K the kernel size, S the stride, and U the number of units.}
    \label{tab:archfashion}
\end{table}
\begin{table}[]
    \centering
    \begin{tabular}{c|c}\toprule
Layer & Parameters \\\midrule
     Conv2D & F=64, K=5, S=2, ReLU-N \\
     Conv2D & F=96, K=3, S=1, ReLU-N\\ 
     Conv2D & F=128, K=3, S=2, ReLU-N\\ 
     Flatten & \\
     Fully-connected & U=128, ReLU-N\\
     Fully-connected & U=10\\\bottomrule
\end{tabular}
    \caption{Convolutional neural network architecture used for the CIFAR-10 experiment. F represents the number of filters, K the kernel size, S the stride, and U the number of units.}
    \label{tab:archcifar}
\end{table}

\end{document}